\documentclass[11pt,a4paper]{article}
\usepackage{emnlp2018}
\usepackage{times}
\usepackage{latexsym}
\usepackage{multirow}
\usepackage{cleveref}
\hypersetup{draft} 

\usepackage{url}
\usepackage{amsfonts}
\usepackage{amsmath}
\usepackage{amssymb}
\usepackage{graphicx}
\usepackage[boxed]{algorithm2e}
\usepackage{subcaption}
\usepackage{booktabs}
\usepackage{bbm}

\usepackage{xcolor}
\usepackage{todonotes}
\usepackage[normalem]{ulem}
\newcommand{\ChM}[2][]{%
\textcolor{black}{#2}}

\newcommand{\YG}[2][]{\sout{\textcolor{red}{#1}}\textcolor{black}{#2}}

\aclfinalcopy 
\def\aclpaperid{1293} 

\crefformat{section}{\S#2#1#3} 
\crefformat{subsection}{\S#2#1#3}
\crefformat{subsubsection}{\S#2#1#3}

\title{APRIL: Interactively Learning to Summarise by Combining
\\ Active Preference Learning and Reinforcement Learning }

\author{Yang Gao,  Christian M. Meyer,  Iryna Gurevych \\
Ubiquitous Knowledge Processing Lab (UKP-TUDA) \\
Department of Computer Science, Technische Universit\"at Darmstadt \\
{\tt https://www.ukp.tu-darmstadt.de/}
\\}

\date{}

\begin{document}
\maketitle
\begin{abstract}
We propose a method to perform automatic document summarisation 
without using reference summaries.
Instead, our method interactively learns from users' \emph{preferences}.
The merit of preference-based interactive summarisation is that
preferences are easier for users to provide than reference summaries.
Existing preference-based interactive learning methods
suffer from high sample complexity, i.e.\ they need to interact with
the oracle for many rounds in order to converge. 
In this work, we propose a new objective function,
which enables us to leverage active learning, preference learning
and reinforcement learning techniques in order to reduce the sample complexity.
Both simulation and real-user experiments suggest
that our method significantly advances the state of the art.
Our source code is freely available at 
\url{https://github.com/UKPLab/emnlp2018-april}.
\end{abstract}

\section{Introduction}
With the rapid growth of text-based information 
on the Internet, \emph{automatic document summarisation} 
attracts increasing research attention 
from the Natural Language Processing (NLP) community 
\cite{nenkova2012survey}.
Most existing document summarisation techniques require
access to reference summaries to train their systems.
However, obtaining reference summaries is very expensive:
\citet{lin2004rouge} reported that 3,000 hours of human effort 
were required for a simple evaluation of the summaries 
for the Document Understanding Conferences (DUC).
Although previous work has proposed heuristics-based methods to
summarise without reference summaries 
\cite{ryang2012emnlp,rioux2014emnlp}, the gap between
their performance and the upper bound is still large:
the ROUGE-2 upper bound of .212 on DUC'04 \cite{avinesh_meyer2017_interactive_doc_sum} 
is, for example, twice as high as \citeauthor{rioux2014emnlp}'s (\citeyear{rioux2014emnlp}) .114.

\begin{figure}
\begin{subfigure}[b]{0.5\textwidth}
\centering
\includegraphics[width=0.5\textwidth]{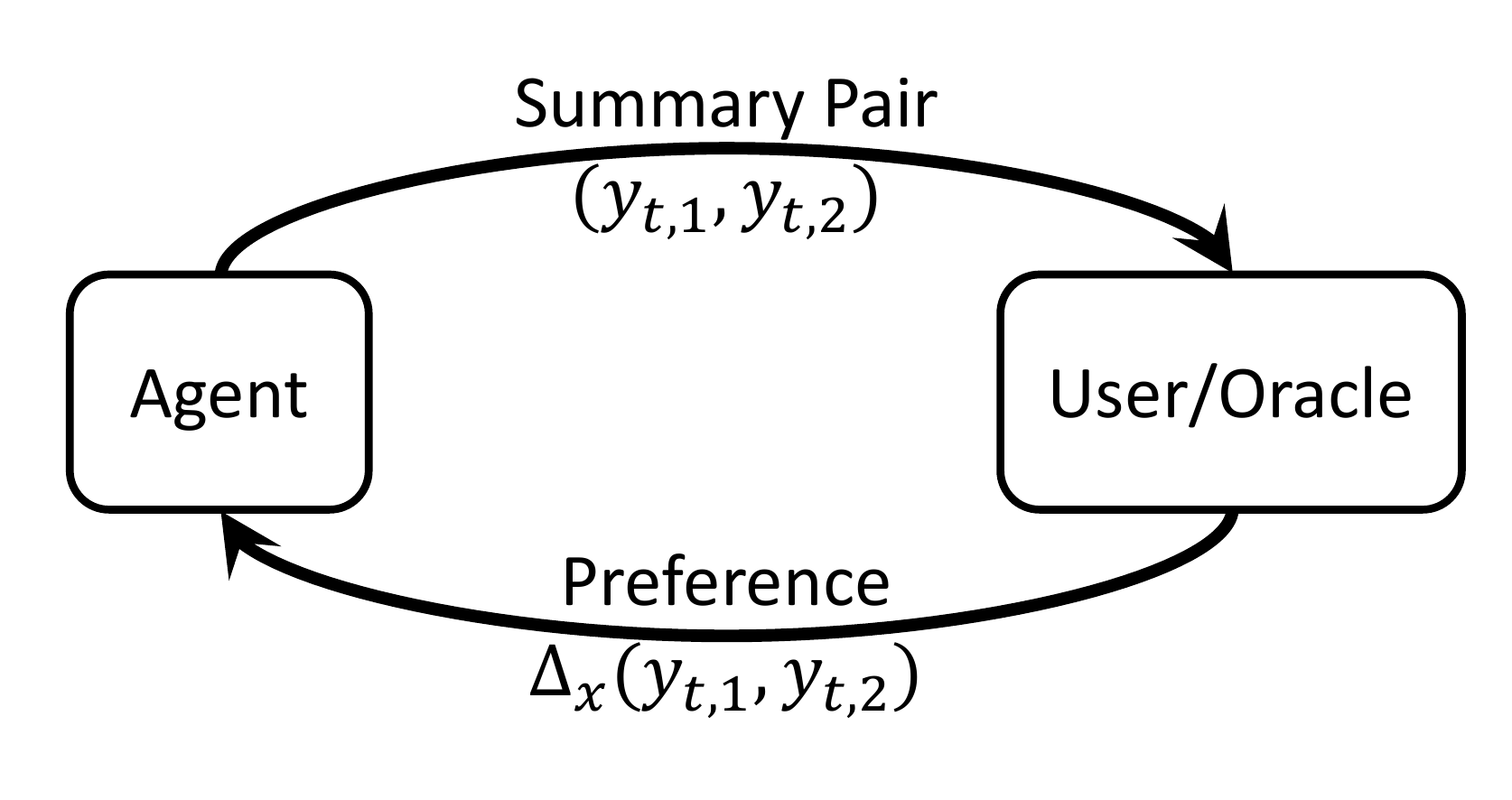}
\caption{Workflow of preference-based SPPI}
\label{subfig:sppi_workflow}
\end{subfigure}
\\
\\
\begin{subfigure}[b]{0.5\textwidth}
\centering
\includegraphics[width=0.95\textwidth]{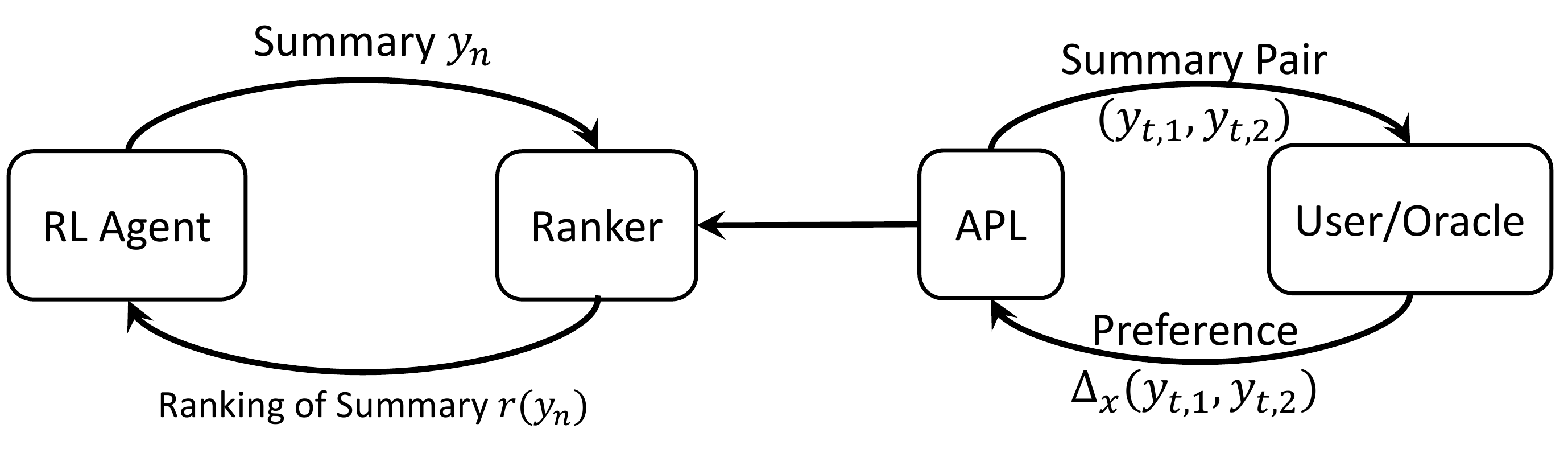}
\caption{Workflow of APRIL} 
\label{subfig:our_workflow}
\end{subfigure}
\caption{A comparison of workflows of SPPI (a) and APRIL (b)
in the EMDS use case. \ChM[Details of the notations (]{Notation details,} e.g., $\Delta_x$ and $r(y_n)$\ChM[)]{,} 
are \ChM{discussed} in \cref{sec:backgrond}.}
\label{fig:compare_workflow}
\end{figure}

The \emph{Structured Prediction from Partial Information} (SPPI) framework
\ChM[is]{has been} proposed to learn to make structured predictions
without access to gold standard data 
\cite{DBLP:conf/nips/SokolovKRL16}.
SPPI is an interactive NLP paradigm: It interacts with a user for
multiple rounds and learns from the user's feedback.
SPPI can learn from two forms of feedback:
\emph{point-based} feedback, i.e. a numeric score for the presented prediction,
or \emph{preference-based} feedback, i.e.\ a preference over a pair of predictions. 
Providing preference-based 
feedback yields a lower cognitive burden for humans than 
providing ratings or categorical labels 
\cite{thurstone27comparitive,kendall1948rank,kinsley2010perference,conf/naacl/zopf18pref}.
%
%
%
Preference-based SPPI has been applied to multiple NLP applications,
including text classification, chunking and machine translation
\cite{sokolov_etal16interactive_nlp,DBLP:conf/acl/KreutzerSR17}.
However, SPPI has prohibitively high sample complexities
in the aforementioned NLP tasks, as it needs at least hundreds of thousands 
rounds of interaction to make near-optimal predictions, 
even with simulated ``perfect'' users.
Figure \ref{subfig:sppi_workflow} illustrates
the workflow of the preference-based SPPI.

To reduce the sample complexity, in this work, we propose a novel
preference-based interactive learning framework, 
called \emph{APRIL} (Active Preference ReInforcement Learning).
APRIL goes beyond SPPI by proposing a new objective function,
which divides the preference-based interactive 
learning problem into two phases
(illustrated in Figure \ref{subfig:our_workflow}):
an \emph{Active Preference Learning} (APL) phase (the right
cycle in Figure \ref{subfig:our_workflow}), 
and a \emph{Reinforcement Learning} (RL) phase (the left cycle).
We show that this separation enables us to 
query preferences more effectively and to use the collected
preferences more efficiently, so as to reduce the sample complexity.

We apply APRIL to
\emph{Extractive Multi-Document Summarisation} 
(EMDS). The task of EMDS is to extract
sentences from the original documents to build 
a summary under a length constraint.
We accommodate multiple APL and RL techniques
in APRIL and compare their performance
under different simulation settings.
We also compare APRIL to a state-of-the-art SPPI implementation
using both automatic metrics and human evaluation.
Our results suggest that APRIL significantly outperforms SPPI.

\section{Related Work}
\label{sec:related_work}

RL
has been \ChM{previously }used to perform EMDS without using reference summaries.
\citet{ryang2012emnlp} formulated EMDS as a \emph{Markov Decision
Process} (MDP), designed a heuristics-based reward function 
considering both information coverage rate and redundancy level,
and used the \emph{Temporal Difference} (TD) algorithm 
\cite{Sutton84} to solve the MDP.
In a follow-up work, 
\citet{rioux2014emnlp} proposed a different reward function,
which also did not require reference summaries;
their experiments suggested that using their new reward 
function improved the summary quality.
\YG{\citet{ig2015rl} proposed a different RL formulation of EMDS and 
jointly used supervised learning and RL to perform the task.
However, their method requires the access to reference summaries.}
More recent works applied encoder-decoder-based RL
to document summarisation \cite{DBLP:journals/corr/RanzatoCAZ15,DBLP:conf/naacl/NarayanCL18,DBLP:journals/corr/PaulusXS17,pasunuru2018naacl_rl_summ}. 
These works outperformed standard encoder-decoder as RL can 
directly optimise the ROUGE scores
and can tackle the \emph{exposure bias} problems.
However, these neural RL methods all used ROUGE scores as their
rewards, which in turn relied on reference summaries.
APRIL can accommodate these neural RL techniques
in its RL phase by using a ranking of summaries instead
of the ROUGE scores as rewards. 
We leave neural APRIL for future study.

\citet{avinesh_meyer2017_interactive_doc_sum} proposed a bigram-based
interactive EMDS framework. They asked users
to label important bigrams in candidate summaries and used
\emph{integer linear programming} (ILP) to extract sentences
covering as many important bigrams as possible.
Their method requires no access to reference summaries, 
but it requires considerable human effort during
the interaction: in simulation experiments, their system
needed to collect up to 350 bigram annotations from a (simulated) user.
In addition, they did not consider noise in users' annotations 
but simulated perfect oracles.

\emph{Preference learning} \ChM[studies]{aims at} obtaining the ranking (i.e.\ total
ordering) of objects from pairwise preferences
\cite{DBLP:books/daglib/p/FurnkranzH10}.
\citet{edwin2018tacl} proposed to use an improved \emph{Gaussian process
preference learning} \cite{DBLP:conf/icml/ChuG05} 
for learning to rank arguments in terms of convincingness
from crowdsourced annotations. However, such Bayesian methods
can hardly scale and suffer from high computation time.
\citet{conf/naacl/zopf18pref} recently
proposed to learn a sentence ranker from preferences. 
The resulting ranker can be used to identify the important
sentences and thus to evaluate the quality of the summaries. 
His study also suggests that providing sentence preferences
takes less time than writing reference summaries.
APRIL not only learns a ranking
over summaries from pairwise preferences, but also 
uses the ranking to ``guide'' our RL agent to 
generate good summaries.

There is a recent trend in machine learning
to combine active learning, preference learning and RL,
for learning to perform complex tasks from preferences
\cite{DBLP:journals/jmlr/WirthANF17}. The resulting algorithm
is termed \emph{Preference-based RL} (PbRL), and has been used in
multiple applications, including training robots 
\cite{wirth_etal2016aaai_pbrl} and Atari-playing agents \cite{christiano2017pbrl}.
SPPI and APRIL can both be viewed as PbRL algorithms.
But unlike most PbRL methods that learn 
a utility function of the predictions
(in EMDS, predictions are summaries) to guide the RL agent, 
APRIL is able to directly use a ranking of predictions
to guide the RL agent without making assumptions
about the underlying structure of the utility functions.
This also enables APRIL to use non-utility-based 
preference learning techniques
(e.g.,\ \citealp{DBLP:conf/icml/MaystreG17}).

\section{Background}
\label{sec:backgrond}
In this section, we recap necessary details of SPPI, RL
and preference learning, and adapt them to the EMDS use case,
laying the foundation for APRIL.

\subsection{The SPPI Framework}
\label{subsec:sppi}
Let $\mathcal{X}$ be the input space and let $\mathcal{Y}(x)$
be the set of possible outputs for input $x \in \mathcal{X}$.
In EMDS, $x \in \mathcal{X}$ is a cluster of documents and
$\mathcal{Y}(x)$ is the set of all possible summaries for cluster $x$.
The function $\Delta_x\colon \mathcal{Y}(x) \times \mathcal{Y}(x) \to \{0,1\}$ is the 
\emph{preference function}
such that $\Delta_x(y_i,y_j) = 1$ if
the user believes $y_j$ is better than $y_i$ (denoted by $y_j \succ y_i$
or equivalently $y_i \prec y_j$), and 0 otherwise.
Throughout this paper we assume that users do not equally prefer two different items.
For a given $x$, the expected loss is:
\begin{align}
& \mathcal{L}^\mathrm{SPPI}(w|x) = \mathbb{E}_{p_w(y_i,y_j|x)}[
\Delta_x(y_i,y_j)] \nonumber \\ 
& =  
\sum_{y_i,y_j \in \mathcal{Y}(x)} \Delta_x( y_i,y_j) \;
p_w( y_i,y_j |x) \label{eq:sr_loss},
\end{align}
where $p_w( y_i,y_j |x)$ is the probability of querying the
pair $( y_i,y_j )$. Formally,
\begin{align}
& p_w( y_i,y_j |x) \nonumber \\
& = \frac{\exp[w^\intercal  (\phi(y_i|x)-\phi(y_j|x))]}
{\sum\limits_{y_p,y_q \in \mathcal{Y}(x)}
\exp[w^\intercal  (\phi(y_p|x) - \phi(y_q|x))]} \label{eq:sr_query},
\end{align}
where $\phi(y|x)$ is the vector representation of $y$ given $x$,
and $w$ is the weight vector to be learnt.
Eq.~\eqref{eq:sr_query} is a Gibbs sampling strategy:
$w^\intercal (\phi(y_i|x)-\phi(y_j|x))$ can be viewed as the
``utility gap'' between $y_i$ and $y_j$. The sampling strategy 
$p_w$ encourages querying pairs with large utility gaps.

To minimise $\mathcal{L}^\mathrm{SPPI}$, SPPI uses gradient descent
to update $w$ incrementally. Alg.\ \ref{alg:sr} presents
the pseudo code of our adaptation of SPPI to EMDS.
In the supplementary material, we provide a detailed derivation of $\nabla_w \mathcal{L}^\mathrm{SPPI}(w|x)$.


\begin{algorithm}[t]
\SetKwInOut{Input}{Input}
\SetKwInOut{Output}{Output}
  \Input{sequence of learning rates $\gamma_t$; query budget $T$; 
  document cluster $x$}
	initialise $w_0$\;
  \While{$t=0 \ldots T$}{
  	sampl\ChM[ing]{e} $( y_i,y_j )$ according to Eq.~\eqref{eq:sr_query}\;
    obtain feedback $\Delta_x( y_i,y_j )$\;
    $w_{t+1} := w_t - \gamma_t \nabla_w \mathcal{L}^\mathrm{SPPI}(w|x)$
    }
  \Output{$y^* = \arg \max_{y\in Y(x)} w_{T+1}^\intercal \phi(y,x)$}
  \caption{SPPI for preference-based interactive document summarisation
  (adjusted from Alg.~2 in \cite{sokolov_etal16interactive_nlp}).
  \vspace*{-0.5cm}
  \label{alg:sr}}
\end{algorithm}

\subsection{Reinforcement Learning}
\label{subsec:rl}
RL amounts to efficient algorithms for searching optimal solutions
in MDPs. MDPs are widely used
to formulate \emph{sequential decision making} problems,
which EMDS falls into: in EMDS, the summariser has to sequentially select
sentences from \ChM{the }original documents \ChM[to add]{and add them} to the draft summary.
An (episodic) MDP is a tuple $(S,A,P,R,T)$. $S$ is the set of 
\emph{states}, $A$ is the set of \emph{actions}, 
$P\colon S \times A \times S \to \mathbb{R}$ 
is the \emph{transition function} with $P(s'|s,a)$ yielding 
the probability of performing action $a$
in state $s$ and being transited to a new state $s'$. 
$R\colon S \times A \to \mathbb{R}$
is the \emph{reward function} with $R(s,a)$ giving the immediate reward for
performing action $a$ in state $s$. 
$T \subseteq S$ is the set of 
\emph{terminal states}; visiting a terminal state terminates the
current episode. 

In EMDS, we follow the same MDP formulation
as \citet{ryang2012emnlp} and \citet{rioux2014emnlp}. Given a document 
cluster, a state $s$ is a draft summary, $A$ includes two types of 
actions, \emph{concatenate} a new sentence to the current draft summary,
or \emph{terminate} the draft summary construction.
The transition function $P$ in EMDS is trivial because 
given the current draft summary
and an action, the next state can be easily inferred. 
The reward function $R$ returns an evaluation score of the summary
once the action \emph{terminate} is performed; otherwise it returns 0
because the summary is still under construction and thus not ready
to be evaluated. Providing non-zero rewards before the action \emph{terminate}
can lead to even worse result, as reported by \citet{rioux2014emnlp}.

A \emph{policy} $\pi\colon S \times A \to \mathbb{R}$ in an MDP 
defines how actions are selected: $\pi(s,a)$ is the probability
of selecting action $a$ in state $s$.
In EMDS, a policy corresponds to a strategy to build summaries 
for a given document cluster. We let $\mathcal{Y}_{\pi}(x)$
be the set of all possible summaries the policy $\pi$ can construct
in the document cluster $x$, and we slightly abuse the notation
by letting $\pi(y|x)$ denote the probability of policy $\pi$
generating a summary $y$ in cluster $x$. Then the expected reward
of a policy is:
\begin{align}
\mathcal{R}^\mathrm{RL}(\pi|x) & = \mathbb{E}_{y \in 
\mathcal{Y}_{\pi}(x)} R(y|x) \nonumber\\
& = \sum\limits_{y \in \mathcal{Y}_{\pi}(x)} \pi(y|x)R(y|x),
\label{eq:rl_obj}
\end{align}
where $R(y|x)$ is the reward for summary $y$ in
document cluster $x$.
The goal of an MDP is to find the optimal policy $\pi^*$
that has the highest expected reward: $\pi^* =\arg\max_{\pi}\mathcal{R}^\mathrm{RL}(\pi)$.

Note that the loss function in SPPI (Eq.\ \eqref{eq:sr_loss})
and the expected reward function in RL (Eq.\ \eqref{eq:rl_obj})
are in similar forms: if we view the pair selection
probability $p_w$ in Eq.~\eqref{eq:sr_query} as a 
policy, and view the preference function
$\Delta_x$ in Eq.~\eqref{eq:sr_loss}
as a negative reward function, we can view SPPI as an RL problem.
The major difference between SPPI and RL is that 
SPPI selects and evaluates pairs
of outputs, while RL selects and evaluates single outputs.
We will exploit their connection
to propose our new objective function and the APRIL framework.

\subsection{Preference Learning}
\label{subsec:apl}
The linear Bradley-Terry (BT) model \cite{bradley1952rank} is one of 
the most widely used methods in preference learning.
Given a set of items $\mathcal{Y}$,
suppose we have observed $T$ preferences: 
$Q = \{q_1(y_{1,1},y_{1,2}), \cdots, q_T(y_{T,1},y_{T,2})\}$, 
where $y_{i,1}, y_{i,2} \in \mathcal{Y}$, 
and $q_i \in \{\prec, \succ\}$ is the oracle's preference 
in the $i^\mathrm{th}$ round.
The BT model minimises the following cross-entropy loss:
\begin{align}
\mathcal{L}^\mathrm{BT}(w)= - &\: \hspace*{-.6cm} \sum_{q_i(y_{i,1}, y_{i,2}) \in Q}  
     \hspace*{-.6cm} \quad[\;  \mu_{i,1} \,\log \mathcal{P}_w(y_{i,1} \succ y_{i,2}) \nonumber \\ 
      & +\;  \mu_{i,2} \,\log \mathcal{P}_w(y_{i,2} \succ y_{i,1}) \, ],
      \label{eq:mal_bt}
\end{align}
where $\mathcal{P}_w(y_i \succ y_j) = 
(1+\exp[w^\intercal (\phi(y_j)-\phi(y_i))])^{-1}$, and 
$\mu_{i,1}$ and $\mu_{i,2}$ indicate the direction of preferences:
if $y_{i,1} \succ y_{i,2}$ then $\mu_{i,1} = 1$ and $\mu_{i,2} = 0$.
Let $w^* = \arg \min_w \mathcal{L}^\mathrm{BT}(w)$, then $w^*$ can be used
to rank all items in $\mathcal{Y}$: for any $y_i, y_j \in \mathcal{Y}$,
the ranker prefers $y_i$ over $y_j$ if $w^{*\intercal}\phi(y_i) >
w^{*\intercal}\phi(y_j)$.

\section{APRIL: Decomposing SPPI into Active Preference Learning and RL}
\label{sec:our_method}
A major problem of SPPI is its high sample complexity.
We believe this is due to two reasons.
First, SPPI's sampling strategy is inefficient:
From Eq.\ \eqref{eq:sr_query} we can 
see that SPPI tends to select pairs with large quality gaps
for querying the user. This strategy can quickly identify
the relatively good and relatively bad summaries, but needs many rounds of interaction
to find the top summaries. 
Second, SPPI uses the collected preferences ineffectively:
In 
Alg.\ \ref{alg:sr}, 
each preference is used only once for performing
the gradient descent update and is forgotten afterwards.
SPPI does not generalise or re-use collected preferences, 
wasting the useful and expensive information.

These two weaknesses of SPPI motivate us to propose a new
learning paradigm that can query and generalise
preferences more efficiently\ChM[, being defined as follows]{}.
Recall that in EMDS, the goal is to find the optimal summary for a given
document cluster $x$, namely the summary that is preferred over all
other possible summaries in $\mathcal{Y}(x)$. Based on this understanding,
we define a new expected reward function $\mathcal{R}^\mathrm{APRIL}$ 
for policy $\pi$ as follows:
\begin{align}
& \mathcal{R}^\mathrm{APRIL}(\pi|x) \! = \!\mathbb{E}_{y_j \sim \pi} [
\frac{1}{|\mathcal{Y}(x)|} \hspace{-0.1cm}
\sum\limits_{y_i\in \mathcal{Y}(x)} \hspace{-0.2cm} \Delta_x(y_i,y_j) ] \nonumber \allowdisplaybreaks
\\
& = \frac{1}{|\mathcal{Y}(x)|} \sum\limits_{y_j\in \mathcal{Y}_\pi(x)}
\pi(y_j|x)
\sum\limits_{y_i\in \mathcal{Y}(x)} \Delta_x(y_i,y_j) 
\nonumber \allowdisplaybreaks
\\
& = \sum\limits_{y\in \mathcal{Y}_\pi(x)} \pi(y|x) \; r(y|x),
\label{eq:our_obj}
\end{align}
where $r(y|x)\! =\! \sum_{y_i\in \mathcal{Y}(x)} 
\Delta_x( y_i,y_j )/|\mathcal{Y}(x)|$.
Note that $\Delta_x( y_i,y_j )$ equals 1 if 
$y_j$ is preferred over $y_i$ and equals 0 otherwise
(see \cref{subsec:sppi}). Thus, $r(y|x)$
is the relative position of $y$ in the (ascending) sorted
$\mathcal{Y}(x)$, and it can be approximated by preference learning.
The use of preference learning enables us to generalise
the observed preferences to a ranker (see \cref{subsec:apl}),
\YG{allowing more effective use of the collected preferences.}
Also, we can use active learning to 
select summary pairs for querying more effectively. 
In addition, the resemblance of $\mathcal{R}^\mathrm{APRIL}$ 
and RL's reward function $\mathcal{R}^\mathrm{RL}$ (in Eq. \eqref{eq:rl_obj})
enables us to use a wide range of RL algorithms
to maximise $\mathcal{R}^\mathrm{APRIL}$
(see \cref{sec:related_work}).

Based on the new objective function,  we split the 
preference-based interactive learning into two phases:
an \emph{Active Preference Learning} (APL) phase (the right cycle in Fig.\
\ref{subfig:our_workflow}), responsible for  querying preferences 
from the oracle and approximating the  
ranking of summaries, and an \emph{RL} phase (the
left cycle in Fig. \ref{subfig:our_workflow}), 
responsible for learning to summarise based on the learned ranking. 
The resulting framework APRIL allows for integrating any active preference
learning and RL techniques.
Note that only the APL phase is online (i.e.\ involving \ChM[people]{humans} in the loop)
while the RL phase can be performed offline, 
helping to improve the real-time responsiveness.
Also, the learned ranker can provide an unlimited number of 
rewards (i.e.\ $r(y|x)$ in Eq.\ \eqref{eq:our_obj}) to the RL agent,
enabling us to perform many episodes of RL training with a small
number of collected preferences -- unlike in SPPI where
each collected preference is used to train the
system for one round and is forgotten afterwards.
%
Alg.~\ref{alg:our} shows APRIL in pseudo code.

\begin{algorithm}[t]
\SetKwInOut{Input}{Input}
\SetKwInOut{Output}{Output}
  \Input{query budget $T$; document cluster $x$; RL episode budget $N$}
  /* Phase 1: active preference learning */ \\
  \While{$t=0 \ldots T$}{
  	sample a summary pair $( y_i, y_j )$  using \ChM[some]{any} APL strategy\;
    obtain feedback $\Delta_x( y_i, y_j )$\;
    update ranker according to Eq.\ \eqref{eq:mal_bt} \;
    }
  /* Phase 2: RL-based summarisation */ \\
  initialise an arbitrary policy $\pi_0$\;
  \While{$n=0 \ldots N$}{
  	evaluate policy $\pi_n$ according to Eq.\ \eqref{eq:our_obj}\;
    update policy $\pi_n$ using any RL algorithm\;
  }
  \Output{$y^* = \arg \max_{y \in \mathcal{Y}_{\pi_N}(x)} \pi_N(y|x)$}
  \caption{Pseudo code of APRIL
  \label{alg:our}}
\end{algorithm}

\section{\ChM[Simulation ]{}Experimental Setup}
\label{sec:exp_setup}

\paragraph{Datasets.} 
We perform experiments on DUC$\,$'04 to find the best
performing APL and RL techniques. Then we combine
the best-performing APL and RL to complete APRIL and
compare it against SPPI on the 
DUC$\,$'01, DUC$\,$'02 and DUC$\,$'04
datasets.\footnote{\url{http://duc.nist.gov/}}
Some statistics of these datasets are summarised in
Table~\ref{table:datasets}. 

\begin{table}[t]
  \centering\small
    \begin{tabular}{l c c c c}
      \toprule
      Dataset & Lang & \#\,Topic & \#\,Doc & \#\,Token/Doc \\
      \midrule
      DUC$\,$'01 & EN & 30 & 308 & 781 \\
      DUC$\,$'02 & EN & 59 & 567 & 561 \\
      DUC$\,$'04 & EN & 50 & 500 & 587 \\
      \bottomrule
    \end{tabular}
  \caption{Statistics of the datasets. The target summary length
  is 100 tokens in all three datasets.
  \vspace*{-0.5cm}}
  \label{table:datasets}
\end{table}

\paragraph{Simulated Users.}
Existing preference-based interactive learning techniques
assume that the oracle has  an \emph{intrinsic 
evaluation function} $U^*$ and provides preferences consistent
with $U^*$ by preferring higher valued candidates.
We term this a \emph{Perfect Oracle} (PO). We believe
that assuming a PO is unrealistic for real-world applications, because
sometimes real users tend to misjudge the preference direction, especially
when the presented candidates have similar quality.
In this work, besides PO, 
we additionally consider two types of noisy oracles based on
the user-response models proposed by \citet{DBLP:conf/nips/ViappianiB10}:
\begin{itemize}\itemsep0em 
\item \textbf{Constant noisy oracle (CNO)}: with probability $c \in [0,1]$,
this oracle randomly selects which summary is preferred; otherwise
it provides preferences consistent with $U^*$. 
We consider CNOs with $c=0.1$ and $c=0.3$.
\item \textbf{Logistic noisy oracle (LNO)}: 
for two summaries $y_i$ and $y_j$ in cluster $x$, 
the oracle prefers $y_i$ over $y_j$ with probability
$p_{U^*}(y_i \succ y_j | x; m)=(1+\exp[(U^*(y_j|x)-U^*(y_i|x))/m])^{-1}$.
This oracle reflects the intuition that users are more
likely to misjudge the preference direction
when two summaries have similar quality.
Note that the parameter $m \in \mathbb{R}^+$ 
controls the ``noisiness'' of the user's responses: 
higher values of $m$ result in a less steep sigmoid curve,
and the resulting oracle is more likely to misjudge.
We use LNOs with $m=0.3$ and $m=1$.
\end{itemize}
As for the intrinsic evaluation function $U^*$, recent
work has suggested that human preferences over summaries
have high correlations to ROUGE scores 
\cite{conf/naacl/zopf18pref}. Therefore, 
we define: 
\begin{align}
U^*(y|x)\! =\! \frac{R_1(y|x)}{0.47} \!+\! \frac{R_2(y|x)}{0.22} \!+\! 
\frac{R_{S}(y|x)}{0.18}
\label{eq:u_star}
\end{align}
where $R_1$, $R_2$ and $R_{S}$ stand for ROUGE-1, ROUGE-2 and ROUGE-SU4,
respectively. The real values (0.47, 0.22 and 0.18)
are used to balance the weights of the three ROUGE scores.
We choose them to be around the EMDS upper-bound ROUGE scores
reported by \citet{avinesh_meyer2017_interactive_doc_sum}.
As such, an optimal summary's  $U^*$ value should be around 3.

\paragraph{Implementation.}
All code is written in Python and runs on a desktop\ChM{\ PC}
with 8\,GB RAM and an i7-2600 CPU. 
We use NLTK \cite{DBLP:books/daglib/0022921} to perform sentence tokenisation.
\ChM{Our source code is freely available at \url{https://github.com/UKPLab/emnlp2018-april}.}

\section{Simulation Results}
\label{sec:exp_results}
We \ChM[will ]{}first study the \ChM[two phases (i.e. the ]{}APL \ChM{phase (\cref{subsec:apl_results})}
and the RL phase\ChM{\ (\cref{subsec:rl_results})}) separately\ChM[,]{\ by} 
comparing the performance of multiple APL and RL algorithms 
in each phase\ChM[ respectively]{}.
Then, in \cref{subsec:complete_results}, we combine the best performing APL and RL algorithm
to complete Alg.\ \ref{alg:our} and compare \ChM[it]{APRIL} against SPPI.

\subsection{APL Phase Performance}
\label{subsec:apl_results}
Recall that the task of APL is to output a ranking
of all summaries in a cluster. In this subsection,
we test multiple APL techniques and compare
the quality of their resulting rankings.
Two metrics are used: Kendall's $\tau$ \cite{kendall1948rank} and
Spearman's $\rho$ \cite{spearman1904proof}.
Both metrics are valued between $-1$ and $1$, with higher
values suggesting higher rank correlation. 
Because the number of possible summaries in a cluster is huge,
instead of evaluating the ranking quality on all
possible summaries, we evaluate rankings on 
10,000 randomly sampled summaries, denoted $\hat{\mathcal{Y}}(x)$.
During querying, all candidate summaries presented to the oracle 
are also selected from $\hat{\mathcal{Y}}(x)$. 
Sampling $\hat{\mathcal{Y}}(x)$ a priori
helps us to reduce the response time to under 500 ms for
all APL techniques we test. 
We compare four active learning strategies under
two query budgets, $T = 10$ and $T = 100$:



\begin{itemize}
\itemsep0em 
\item \textbf{Random Sampling (RND)}:
Randomly select two summaries from $\hat{\mathcal{Y}}(x)$ to
query.
\item \textbf{SPPI Sampling (SBT)}:
Select summary pairs from $\hat{\mathcal{Y}}(x)$
according to \ChM{the SPPI strategy in }Eq.\ \eqref{eq:sr_query}. After each round,
the weight vector $w$ is updated according to Eq.\ \eqref{eq:mal_bt}.
\item \textbf{Uncertainty Sampling (Unc)}:
Query the most \emph{uncertain} summary pairs.
In line with \citet{avinesh_meyer2017_interactive_doc_sum},
the uncertainty of a summary is evaluated as follows:
first, we estimate the probability of a summary $y$
being the optimal summary in cluster $x$ as
 $ p_\mathrm{opt}(y|x) = 
 (1+\exp(-w_t^{*\intercal}\phi(x,y)))^{-1}$,
where $w_t^*$ is the weights learned by the BT model (see
\cref{subsec:apl}) in round $t$. Given $p_\mathrm{opt}(y|x)$, we let the uncertainty score 
$unc(y|x) = 1-p_\mathrm{opt}(y|x)$ if $p_\mathrm{opt}(y|x) \ge 0.5$ 
and $unc(y|x) = p_\mathrm{opt}(y|x)$ otherwise.
\item \textbf{J\&N}
is the \emph{robust query selection algorithm}
proposed by \citet{DBLP:conf/nips/JamiesonN11}.
It assumes \ChM{that the }items' preferences are dependent on
their distances to an unknown reference point in the 
embedding space: the farther an item to the reference
point, the more preferred the item is. After each round
of interaction, the algorithm uses all collected preferences
to locate the area where the reference point may fall into,
and identify the query pairs which can reduce the size of this
area, termed \emph{ambiguous query pairs}. 
To combat noise in preferences, the algorithm selects the 
most-likely-correct ambiguous pair
to query the oracle in each round.
\end{itemize}

After all preferences are collected, we obtain the ranker
as follows: for any $y_i, y_j \in \mathcal{Y}(x)$,
the ranker prefers $y_i$ over $y_j$ if 
\begin{align}
& \alpha w^{*\intercal}\phi(y_i|x)+(1-\alpha)HU(y_i|x) > &
\nonumber \\
&\alpha w^{*\intercal}\phi(y_j|x)+(1-\alpha) HU(y_j|x), &
\label{ineq:rank}
\end{align}
where $w^*$ is the weights vector learned by the 
BT model (see Eq.\ \eqref{eq:mal_bt}), $HU$ is the 
heuristics-based summary evaluation function proposed 
by \citet{ryang2012emnlp},
and $\alpha \in [0,1]$ is a parameter.
The aim of using $HU$ and $\alpha$ is to trade off between
the \emph{prior knowledge} (i.e.\ heuristics-based $HU$) and the
\emph{posterior observation} (i.e.\ the BT-learnt $w^*$),
so as to combat the \emph{cold-start} problem. 
Based on some preliminary experiments, 
we set $\alpha=0.3$ when the query budget is 10, and $\alpha=0.7$
when the query budget is 100. 
The intuition is to put more weight to the posterior 
with increasing rounds of interaction.
More systematic research of $\alpha$ can yield better results; 
we leave it for future work.
For the vector $\phi(y|x)$, we use the same 
bag-of-bigram embeddings as \citet{rioux2014emnlp}, and
we let its length be 200. 

In Table \ref{table:apl_results}, 
we compare the performance of the four APL methods on the DUC'04 dataset.
The baseline we compared against is the prior ranking.
We find that Unc significantly\footnote{In this paper we use double-tailed 
student t-test to compute p-values, and we let significance level 
be $p<0.01$.} outperforms all other APL methods,
except when the oracle is LNO-1, where the advantage
of Unc to SBT is not significant. Also, both Unc and SBT 
are able to significantly
outperform the baseline under all settings.
The competitive performance of SBT, especially with LNO-1,
is due to its unique sampling strategy: LNO-1 is more likely
to misjudge the preference direction when the presented
summaries have similar quality, but SBT has high
probability to present summaries with large quality gaps 
(see Eq.\ \eqref{eq:sr_query}), effectively reducing the chance that
LNOs misjudge preference directions. 
However, SBT is more ``conservative'' compared to Unc
because it tends to exploit the existing ranking
to select one good and one bad summary to query,
while Unc performs more exploration by querying
the summaries that are least confident according to the current ranking.
We believe this explains the strong overall
performance of Unc. 

%
Additional experiments suggest that
when we only use the posterior ranking (i.e.\
letting $\alpha=1$),
no APL we test can surpass the baseline when $T=10$.
Detailed results are presented in the supplementary material.
This observation reflects the severity 
of the cold-start problem, confirms the effectiveness of our
prior-posterior trade-off mechanism in combating cold-start,
and indicates the importance 
of tuning the $\alpha$ value (see Eq.\ \eqref{ineq:rank}). 
This opens up  exciting avenues for future work.

\begin{table}
\small
\centering
\hspace*{-.1cm}
\begin{tabular}{@{\hspace{.03cm}}l@{\hspace{-.1cm}}*{1}{@{\hspace{.08cm}}} *{4}{  @{\hspace{.29cm}} c @{\hspace{.17cm}} c } @{\hspace{.03cm}}}
\toprule & \multicolumn{2}{c}{RND} & \multicolumn{2}{c}{SBT} 
    & \multicolumn{2}{c}{Unc} & \multicolumn{2}{c}{J\&N} \\
Oracle & $\tau$ & $\rho$ & $\tau$ & $\rho$ & $\tau$ & $\rho$ & $\tau$ & $\rho$ \\
\midrule
\multicolumn{9}{l}{\hspace*{-0.2cm}\textit{Query budget $T =$ 10, $\alpha=0.3$:}} \\
PO & .211 & .310 & .241 & .353 
& \textbf{.253}$^*$ & \textbf{.370}$^*$
& .217 & .319 \\ 
CNO-0.1 & .208 & .307 & .231 & .339 
& \textbf{.240}$^*$ & \textbf{.351}$^*$ 
& .211 & .311 \\
CNO-0.3 & .210 & .309 & .218 & .320 
& \textbf{.229}$^*$ & \textbf{.337}$^*$ 
& .205 & .302 \\
LNO-0.3 & .210 & .309 & .216 & .318 
& \textbf{.231}$^*$ & \textbf{.339}$^*$ 
& .209 & .307 \\
LNO-1 & .206 & .303 & .210 & .308 & \textbf{.211} & \textbf{.310} 
& .207 & .305 \\
\midrule
\multicolumn{9}{l}{\hspace*{-0.2cm}\textit{Query budget $T =$ 100, $\alpha=0.7$:}} \\
PO & .258 & .377 & .340 & .490 
& \textbf{.418}$^*$ & \textbf{.587}$^*$ 
&.255 & .372 \\
CNO-0.1 & .248 & .363 & .317 & .459 
& \textbf{.386}$^*$ & \textbf{.549}$^*$ 
& .247 & .362 \\
CNO-0.3 & .212 &.312 & .271 & .396 
& \textbf{.330}$^*$ & \textbf{.476}$^*$ 
& .232 & .340 \\
LNO-0.3 & .231 & .339 & .277 & .404 
& \textbf{.324}$^*$ & \textbf{.467}$^*$ 
&.229 & .336 \\
LNO-1 & .210 & .309 & \textbf{.225} & .330 
& \textbf{.225} & \textbf{.331} & .213 & .313  \\
\midrule
\multicolumn{9}{l}{\hspace*{-0.2cm} \emph{Baseline,} $\alpha=0$, $T=0$: 
$\tau=.206$, $\rho=.304$ } \\
\bottomrule
\end{tabular}
\caption{Performance of multiple APL algorithms (columns) using different oracles and query budgets (rows). 
The baseline is the purely prior ranking.
All results except the baseline are averaged 
over 50 document clusters in DUC'04.
Asterisk: significant advantage over other active learning strategies
given the same oracle and budget $T$.} 
\vspace*{-0.5cm}
\label{table:apl_results}
\end{table}

\subsection{RL Phase Performance}
\label{subsec:rl_results}
We compare two RL algorithms: TD($\lambda$) \cite{Sutton84}
and LSTD($\lambda$) \cite{DBLP:conf/icml/Boyan99}. 
TD($\lambda$) has been used
in previous RL-based EMDS work \cite{ryang2012emnlp,rioux2014emnlp}.
LSTD($\lambda$) is chosen, because it is an improved TD algorithm
and has been used 
in the state-of-the-art PbRL algorithm by \citet{wirth_etal2016aaai_pbrl}.
We let the learning round (see Alg.\ \ref{alg:our}) $N = 5,000$, 
which we found to yield good results in reasonable time 
(less than 1 minute to generate a summary for one document cluster).
Letting $N=3,000$ will result in a significant performance drop,  while
increasing $N$ to 10,000 will only bring marginal improvement
at the cost of doubling the runtime.
The learning parameters we use
for TD($\lambda$) are the same as those by
\citet{rioux2014emnlp}. For LSTD($\lambda$), we 
let $\lambda=1$ and initialise its square matrix as a 
diagonal matrix with random numbers between 0 
and 1, as suggested by \citet{lagoudakis2003lspi}. 
The rewards we use are
the $U^*$ function introduced in \cref{sec:exp_setup}.
Note that this serves as the upper-bound performance, because
$U^*$ relies on the reference summaries
(see Eq.\ \eqref{eq:u_star}), which are not available
in the interactive setting. 
As a baseline, we also present the upper-bound performance
of integer linear programming (ILP) reported by \citet{avinesh_meyer2017_interactive_doc_sum},
optimised for bigram coverage.

Table \ref{table:rl_perf} shows the performance of RL 
and ILP on the DUC'04 dataset.
TD($\lambda$) significantly outperforms LSTD($\lambda$) 
in terms of all ROUGE scores we consider. 
Although the least-square RL algorithms (which LSTD belongs to)
have been proved to achieve better performance than standard
TD methods in large-scale problems (see \citealp{lagoudakis2003lspi}),
their performance is sensitive to many factors, e.g.,
initialisation values in the diagonal matrix, 
regularisation parameters, etc.
We note that a similar observation about the inferior performance
of least-square RL in EMDS is reported by \citet{rioux2014emnlp}.

TD($\lambda$) also significantly outperforms
ILP in terms of all metrics except ROUGE-2. This is not surprising,
because\ChM{\ the bigram-based} ILP is optimised for ROUGE-2, whereas
our reward function $U^*$ considers other metrics as well\ChM{\ (see Eq.\ \eqref{eq:u_star})}.
Since ILP is widely used as a strong baseline for EMDS,
these results confirm the advantage of using RL for EMDS problems.

\begin{table}
  \centering\small
    \begin{tabular}{l  c c c c}
      \toprule
       Method &  $R_1$ & $R_2$ & $R_L$ & $R_{SU4}$  \\
      \midrule
      TD($\lambda$) & \textbf{.484} 
      & .184 & \textbf{.388} & \textbf{.199} \\
      LSTD($\lambda$) & .458 &.159 & .366 & .185  \\
      ILP & .470 & \textbf{.212} & \,N/A & .185 \\
      \bottomrule
    \end{tabular}
  \caption{Upper-bound performance comparison. 
  Results are averaged over all clusters
  in DUC'04.
  }
  \vspace*{-0.5cm}
  \label{table:rl_perf}
\end{table}

\begin{table*}[ht]
\small
\centering
\begin{tabular}{l l @{\hspace{.5cm}} c *{3}{ @{\hspace{.50cm}} l @{\hspace{.23cm}} l @{\hspace{.23cm}} l @{\hspace{.23cm}} l }}
    \toprule & & & \multicolumn{4}{c}{DUC$\,$'01} & \multicolumn{4}{c}{DUC$\,$'02} 
    & \multicolumn{4}{c}{DUC$\,$'04} \\
    Oracle & Method & $T$ & $R_1$ & $R_2$ & $R_L$ & $R_{SU4}$  & $R_1$ & $R_2$ & $R_L$ & $R_{SU4}$  & $R_1$ & $R_2$ & $R_L$ & $R_{SU4}$ \\
    \midrule
   \multirow{4}{*}{PO} &
   SPPI & 10 & .332 & .075 & .264 & .104 &
   .357 & .083 & .280$^\dagger$ & .116
   & .378 & .098 & .299 & .129 \\
   & APRIL & 10 & \textbf{.357} & .087 & \textbf{.283} & \textbf{.119} &
   \textbf{.390} & \textbf{.108} & \textbf{.306} & \textbf{.133} &
   \textbf{.410} & \textbf{.116} & \textbf{.325} & \textbf{.149} \\
   & SPPI & 100 & .353 & .091 & .284 & .119 &
   .391 & .104 & .306 & .136 &
   .392 & .106 & .312 & .140 \\
    & APRIL & 100 & \textbf{.363} & \textbf{.091} & \textbf{.283} & \textbf{.118} &
    \textbf{.393} & \textbf{.107} & \textbf{.310} & \textbf{.137} &
   \textbf{.415} & \textbf{.118} & \textbf{.325} & \textbf{.151} \\
   \midrule
   \multirow{4}{*}{CNO-0.1} &
   SPPI & 10 & .331 & .081 & .265 & .103 &
   .358 & .081 & .279$^\dagger$ & .114$^\dagger$ &
   .372$^\dagger$ & .093$^\dagger$ & .295$^\dagger$ & .125$^\dagger$ \\
   & APRIL & 10 & \textbf{.351} & .081 & \textbf{.276} & .112 &
   \textbf{.376} & \textbf{.102} & \textbf{.296} & \textbf{.126} &
   \textbf{.403} & \textbf{.111} & \textbf{.320} & \textbf{.145} \\
   & SPPI & 100 & .350 & .089 & .279 & .117 &
   .377 & .100 & .294 & .129 &
   .390 & .107 & .309 & .138 \\
   & APRIL & 100 & \textbf{.353} & .084 & \textbf{.280} & \textbf{.115} &
   \textbf{.385} & \textbf{.103} & \textbf{.302} & \textbf{.134} &
   \textbf{.411} & \textbf{.117} & \textbf{.325} & \textbf{.151} \\
   \midrule
   \multirow{4}{*}{CNO-0.3} &
   SPPI & 10 & .320$^\dagger$ & .063$^\dagger$ & .253$^\dagger$ & .096$^\dagger$ &
   .354$^\dagger$ & .080 & .278$^\dagger$ & .113$^\dagger$ &
   .370$^\dagger$ & .093$^\dagger$ & .295$^\dagger$ & 125$^\dagger$ \\
   & APRIL & 10 & .339 & \textbf{.076} & \textbf{.266} & \textbf{.108} &
   \textbf{.370} & \textbf{.091} & .290 & \textbf{.124} &
   \textbf{.394} & \textbf{.104} & \textbf{.312} & \textbf{.138} \\
   & SPPI & 100 & .345 & .079 & .270 & .111 &
   .373 & .094 & .295 & .125 &
   .386 & .104 & .307 & .136 \\
   & APRIL & 100 & \textbf{.349} & \textbf{.081} & \textbf{.275} & \textbf{.109} &
   \textbf{.376} & \textbf{.097} & \textbf{.296} & \textbf{.127} &
   \textbf{.404} & \textbf{.114} & \textbf{.320} & \textbf{.146} \\
   \midrule
   \multirow{4}{*}{LNO-0.3} &
   SPPI & 10 & .319$^\dagger$ & .067$^\dagger$ & .253$^\dagger$ & .096$^\dagger$ &
   .354$^\dagger$ & .083 & .280$^\dagger$ & .113$^\dagger$ &
   .375$^\dagger$ & .095$^\dagger$ & .294$^\dagger$ & .127$^\dagger$ \\
   & APRIL & 10 & \textbf{.347} & \textbf{.084} & \textbf{.275} & \textbf{.109} &
   \textbf{.370} & \textbf{.095} & \textbf{.289} & \textbf{.125} &
   \textbf{.398} & \textbf{.108} & \textbf{.311} & \textbf{.141} \\
   & SPPI & 100 & .321$^\dagger$ & .068$^\dagger$ & .252$^\dagger$ & .097$^\dagger$ &
   .352$^\dagger$ & .080 & .278$^\dagger$ & .112$^\dagger$ &
   .387 & .104 & .309 & .136 \\
   & APRIL & 100 & \textbf{.350} & \textbf{.086} & \textbf{.277} & \textbf{.123} &
   \textbf{.380} & \textbf{.079} & \textbf{.296} & \textbf{.129} &
   \textbf{.407} & \textbf{.112} & \textbf{.321} & \textbf{.147} \\
   \midrule
   \multirow{4}{*}{LNO-1} &
   SPPI & 10 & .314$^\dagger$ & .058$^\dagger$ & .250$^\dagger$ & .092$^\dagger$ &
   .348$^\dagger$ & .076$^\dagger$ & .273$^\dagger$ & .110$^\dagger$ &
   .373$^\dagger$ & .096$^\dagger$ & .297$^\dagger$ & .126$^\dagger$ \\
   & APRIL & 10 & \textbf{.337} & \textbf{.072} & \textbf{.266} & \textbf{.104} &
   \textbf{.362} & \textbf{.085} & \textbf{.286} & \textbf{.119} &
   \textbf{.388} & \textbf{.102} & \textbf{.307} & \textbf{.134} \\
   & SPPI & 100 & .320$^\dagger$ & .064$^\dagger$ & .255$^\dagger$ & .097$^\dagger$ &
   .351$^\dagger$ & .078$^\dagger$ & .273$^\dagger$ & .113$^\dagger$ &
   .381 & .099 & .301 & .132 \\
   & APRIL & 100 & \textbf{.347} & \textbf{.080} & \textbf{.274} & \textbf{.109} &
   \textbf{.369} & \textbf{.089} & \textbf{.286} & \textbf{.123} &
   \textbf{.391} & \textbf{.101} & \textbf{.308} & \textbf{.136} \\
   \midrule
   \multirow{2}{*}{Baselines}
   & SPPI & 0 & .323 & .068 & .259 & .098 & .350 & .077 & .278 & .112
   & .372 & .093 & .293 & .125 \\
   & TD($\lambda$) & 0 & .324 & .069 & .256 & .099 & .350 & .081 & .276 & .113
   & .372 & .086 & .292 & .122 \\
   \bottomrule
\end{tabular}
\caption{Comparison of APRIL and SPPI. 
All results are averaged over all clusters in each dataset. 
Baselines: $HU$-trained SPPI and TD($\lambda$), 
without any interaction (i.e. $T=0$).
Boldface: Comparable (i.e. no significant gaps exist) or significantly
better than SPPI with 100 rounds of interaction, under the same oracle. 
Superscript $\dagger$:
Comparable or significantly worse than the corresponding baseline.}
\label{table:overall_results}
\end{table*}

\subsection{Complete Pipeline Performance}
\label{subsec:complete_results}
Finally, we combine the best techniques of 
the APL and RL phase (namely Unc and 
TD($\lambda$), respectively) to complete APRIL, 
and compare it against SPPI.
As a baseline, we use the heuristic-based 
rewards $HU$ to train both TD($\lambda$) (ranking-based
training, i.e.\ using $HU$ to produce $r(y|x)$ in Eq.\ \eqref{eq:our_obj}
to train) and SPPI (preference-based training, i.e.\
using $HU$ for generating pairs to train SPPI) for 
up to 5,000 episodes. 
The baseline results are presented 
in the bottom rows of Table \ref{table:overall_results}.

\begin{table}
\centering
\small
\begin{tabular}{l c c c c}
\toprule
 & DUC'01 & DUC'02 & DUC'04 & Overall \\
 \midrule
APRIL & \textbf{3.57$\pm$.30} & \textbf{4.14$\pm$.14} & \textbf{3.86$\pm$.40} 
& \textbf{3.86$\pm$.17} \\
SPPI & 2.29$\pm$.29 & 2.14$\pm$.14 & 3.14$\pm$.34 & 2.52$\pm$.18 \\
\bottomrule
\end{tabular}
\caption{Human ratings \ChM[on]{for} the summaries generated by APRIL and SPPI
(mean$\pm$standard error).
\label{table:user_study}}
\vspace*{-0.7cm}
\end{table}

We make the following observations from Table \ref{table:overall_results}.
\textbf{(i)} Given the same oracle, 
the performance of APRIL with 10 rounds of interaction is comparable
or even superior than
that of SPPI after 100 rounds of interaction (see boldface 
in Table \ref{table:overall_results}),
suggesting the strong advantage of APRIL to reduce sample complexity.
\textbf{(ii)} 
APRIL can significantly improve the baseline
with either 10 or 100 rounds of interaction,
but SPPI's performance can be even worse than 
the baseline\ChM{\ (marked by $^\dagger$ in Table \ref{table:overall_results})},
especially under the high-noise low-budget settings
\ChM{\ (i.e., CNO-0.3, LNO-0.3, and LNO-1 with $T = 10$)}.
This is because SPPI lacks a mechanism to balance
between prior and posterior ranking, while APRIL
can adjust this trade-off by tuning $\alpha$
(Eq.\ \eqref{ineq:rank}). This endows APRIL with 
better noise robustness and lower sample complexity
in high-noise low-budget settings. 
Note that the above observations also hold for
the other two datasets, indicating the consistently
strong performance of APRIL across different datasets.

As for\ChM{\ the} overall runtime, when budget $T=100$, APRIL \ChM[in]{on} average takes
2 minutes to interact with \ChM[one]{an} oracle and output a summary,
while SPPI takes around 15 minutes due to its expensive gradient
descent computation (see \cref{subsec:sppi}).


\section{Human Evaluation}
\label{sec:user_study}

\begin{figure*}
\newcommand{\SummaryBig}[2]{\framebox{%
  \begin{minipage}[t][2.7cm][t]{.483\linewidth}%
    \textbf{#1}: {#2}%
  \end{minipage}%
}}
\newcommand{\Summary}[2]{\framebox{%
  \begin{minipage}[t][2.5cm][t]{.483\linewidth}%
    \textbf{#1}: {#2}%
  \end{minipage}%
}}
\scriptsize
\centering
\SummaryBig{Topic d30042 (DUC'04), SPPI}{
After meeting Libyan leader Moammar Gadhafi in a desert tent, U.N. Secretary-General Kofi Annan said he thinks an arrangement for bringing two suspects to trial in the bombing of a Pan Am airliner could be secured in the ``not too distant future.'' TRIPOLI, Libya (AP) U.N. Secretary-General Kofi Annan arrived in Libya Saturday for talks aimed at bringing to trial two Libyan suspects in the 1988 Pan Am bombing over Lockerbie, Scotland. Secretary General Kofi Annan said Wednesday he was extending his North African tour to include talks with Libyan authorities. Annan's one-day, 2nd graf pvs During his Algerian stay,
}
\SummaryBig{Topic d30042 (DUC'04), APRIL}{
TRIPOLI, Libya (AP) U.N. Secretary-General Kofi Annan arrived in Libya Saturday for talks aimed at bringing to trial two Libyan suspects in the 1988 Pan Am bombing over Lockerbie, Scotland. Annan's one-day visit to meet with Libyan leader Col. Moammar Gadhafi followed reports in the Libyan media that Gadhafi had no authority to hand over the suspects. The 60-year-old Annan is trying to get Libya to go along with a U.S.-British plan to try the two suspects before a panel of Scottish judges in the Netherlands for the Dec. 21, 1988, bombing over Lockerbie, Scotland. Sirte is 400 kilometers (250 miles) east of the Libyan capital Tripoli. During his Algerian stay,
}
\Summary{Topic d117i (DUC'02), SPPI}{
The Booker Prize is sponsored by Booker, an international food and agriculture business. The novel, a story of Scottish low-life narrated largely in Glaswegian dialect, is unlikely to prove a popular choice with booksellers, who have damned all six books shortlisted for the prize as boring, elitist and- worst of all- unsaleable. The shortlist of six for the Pounds 20,000 Booker Prize for fiction, announced yesterday, immediately prompted the question 'Who ? ' Japanese writer Kazuo Ishiguro won the 1989 Booker Prize, Britain's top literary award, for his novel ``The Remains of the Day,'' judges announced Thursday. He didn't win.
}
\Summary{Topic d117i (DUC'02), APRIL}{
Australian novelist Peter Carey was awarded the coveted Booker Prize for fiction Tuesday night for his love story, ``Oscar and Lucinda.'' The Booker Prize is sponsored by Booker, an international food and agriculture business, and administered by The Book Trust. British publishers can submit three new novels by British and Commonwealth writers. Six novels have been nominated for the Booker Prize, Britain's most prestigious fiction award, and bookmakers say the favorite is ``The Remains of the Day'' by Japanese author Kazuo Ishiguro. On the day of the Big Event, Ladbroke, the large British betting agency, posted the final odds.
}
\Summary{Topic d19 (DUC'01), SPPI}{
The issue cuts across partisan lines in the Senate, with Minority Leader Bob Dole (R-Kan.) arguing against the White House position on grounds that including illegal aliens in the census is unfair to American citizens.. Loss of Seats Cited. Shelby's amendment says only that the secretary is to ``make such adjustments in total population figures as may be necessary, using such methods and procedures as the secretary determines feasible and appropriate'' to keep illegal aliens from being counted in congressional reapportionment. ``Some states will lose congressional seats because of illegal aliens,'' Dole argued. But there's nothing simple about it.
}
\Summary{Topic d19 (DUC'01), APRIL}{
In a blow to California and other states with large immigrant populations, the Senate voted Friday to bar the Census Bureau from counting illegal aliens in the 1990 population count. But the Senate already has voted to force the Census Bureau to exclude illegal immigrants in preparing tallies for congressional reapportionment. said that Georgia and Indiana both lost House seats after the 1980 Census, and California and New York- centers of illegal immigration- each gained seats. A majority of the members of the House of Representatives has signaled support. The national head count will be taken April 1, 1990.
}
\caption{Summaries generated by SPPI and APRIL used in the human evaluation experiments.
\label{fig:user_study_summaries}}
\end{figure*}

Finally, we invited real users to compare and evaluate the quality of the 
summaries generated by SPPI and APRIL.
We randomly selected three topics
(d19 from DUC'01, d117i from DUC'02 and
d30042 from DUC'04), and let both SPPI and 
our best-performing APRIL interact with PO
for 10 rounds on these topics.
The resulting 100-word summaries, shown
in Figure \ref{fig:user_study_summaries}, were presented to 
seven users, who had already read two background texts
to familiarize with the topic. The users were asked to provide their
preference on the presented summary pairs and rate the summaries
on a 5-point Likert scale with higher scores for better summaries. 
All users are fluent in English. 

In all \ChM[3]{three} topics, all users prefer the APRIL-generated summaries
over the SPPI-generated summaries.
\ChM[Ratings of the summaries are presented in Table \ref{table:user_study}.]{Table \ref{table:user_study} shows the users' ratings.}
The APRIL-generated summaries \ChM{consistently }receive higher
ratings\ChM[ for all topics]{}. 
These results are consistent with our simulation experiments
and confirm the significant advantage of APRIL over SPPI.

\vspace*{-0.1cm}
\section{Conclusion}
\label{sec:conclusion}
\vspace*{-0.1cm}
We propose a novel preference-based interactive learning formulation
named APRIL (Active Preference ReInforcement Learning), 
which is able to make structured predictions
without referring to the gold standard data.
Instead, APRIL learns from preference-based feedback.
We designed a novel objective function for APRIL,
which naturally split\ChM{s} APRIL into 
an \emph{active preference learning} (APL)
phase and a \emph{reinforcement learning} (RL) phase, 
enabling us to leverage a wide
spectrum of active learning, preference learning and RL algorithms
to maximise the output quality with a limited number of 
interaction rounds.
We applied APRIL to the Extractive Multi-Document Summarisation
(EMDS) problem, simulated the users' preference-giving behaviour 
using multiple user-response models, and 
compared the performance of multiple APL and RL techniques.
Simulation experiments indicated that
APRIL significantly improved the summary quality with
just 10 rounds of interaction (even with high-noise \ChM[users]{oracles}),
and significantly outperformed SPPI
in terms of both sample complexity and noise robustness.
Human evaluation results suggested that real users preferred
the APRIL-generated summaries over the SPPI-generated ones.

We identify two major lines of future work.
On the technical side, 
we plan to employ more advanced APL and 
RL algorithms in APRIL, \ChM[for example some]{such as}
sample-efficient Bayesian-based APL algorithms (e.g., \citealp{edwin2018tacl})
and neural RL algorithms (e.g. \citealp{mnih2015human}) to further
reduce the sample complexity of APRIL.
On the experimental side,
a logical next step is to implement an interactive
user interface for APRIL\ChM[, invite real users
to interact with APRIL and ask them to evaluate
the summaries' quality before and after the interaction.]{\ and conduct a larger evaluation study comparing the summary quality before and after the interaction.}
We also plan to apply APRIL
to more NLP applications, \ChM[e.g.,]{including} machine translation\ChM{, information exploration} and
semantic parsing\ChM[, to test its applicability on
further tasks]{}.


\vspace*{-0.1cm}
\section*{Acknowledgement\ChM{s}}
\vspace*{-0.1cm}
This work has been supported by the German Research Foundation 
as part of the QA-EduInf project (grant GU 798/18-1 and grant RI 803/12-1). 
We thank the researchers and students from TU Darmstadt
who participated in our human evaluation experiments.
We also thank Johannes F{\" u}rnkranz, Christian Wirth 
and the anonymous reviewers for their helpful comments.

\bibliographystyle{acl_natbib}
\bibliography{./general_long}

\end{document}